\documentclass[runningheads]{llncs}
\usepackage[T1]{fontenc}
\usepackage{color}
\usepackage{multirow}
\usepackage[normalem]{ulem}
\useunder{\uline}{\ul}{}
\usepackage{booktabs} 
\usepackage{amsmath,amsfonts}
\usepackage{marvosym}
\usepackage{graphicx}
\begin{document}
\title{Audio-Infused Automatic Image Colorization by Exploiting Audio Scene Semantics}
\titlerunning{Audio-Infused Automatic Image Colorization}
%
\author{Pengcheng Zhao\inst{1} \and
Yanxiang Chen\inst{1}\inst{(}\textsuperscript{\Letter}\inst{)}\and
Yang Zhao \inst{1, 2} \and Zhao Zhang \inst{1}}
\authorrunning{P. Zhao et al.}
%
\institute{School of Computer Science and Information Engineering, Hefei University of Technology, Hefei 230601, China \and
Peng Cheng National Laboratory, Shenzhen 518000, China. \\
\email{chenyx@hfut.edu.cn}\\
}
\maketitle              
\begin{abstract}
Automatic image colorization is inherently an ill-posed problem with uncertainty, which requires an accurate semantic understanding of scenes to estimate reasonable colors for grayscale images. Although recent interaction-based methods have achieved impressive performance, it is still a very difficult task to infer realistic and accurate colors for automatic colorization. To reduce the difficulty of semantic understanding of grayscale scenes, this paper tries to utilize corresponding audio, which naturally contains extra semantic information about the same scene. Specifically, a novel and pluggable audio-infused automatic image colorization (AIAIC) method is proposed, which consists of three stages. First, we take color image semantics as a bridge and pretrain a colorization network guided by color image semantics. Second, the natural co-occurrence of audio and video is utilized to learn the color semantic correlations between audio and visual scenes. Third, the implicit audio semantic representation is fed into the pretrained network to finally realize the audio-guided colorization. The whole process is trained in a self-supervised manner without human annotation. Experiments demonstrate that audio guidance can effectively improve the performance of automatic colorization, especially for some scenes that are difficult to understand only from visual modality.

\keywords{Image colorization \and Audiovisual learning \and Scene semantic guidance.}
\end{abstract}
\section{Introduction}
As a classical computer vision task, image colorization aims to recover plausible chromatic dimensions to grayscale images, which plays an important role in many image processing applications, such as image compression \cite{1}, and restoration of legacy photos and videos \cite{2}. However, predicting the missing color channels from a single luminance channel is essentially an ill-posed problem with uncertainty, i.e., each pixel in the input grayscale image may correspond to multiple colors. Therefore, automatic colorization remains a challenging problem that requires a considerable semantic understanding of the grayscale scene \cite{3,5}.
\begin{figure}[!t]
\centering
\includegraphics[width=0.8\linewidth]{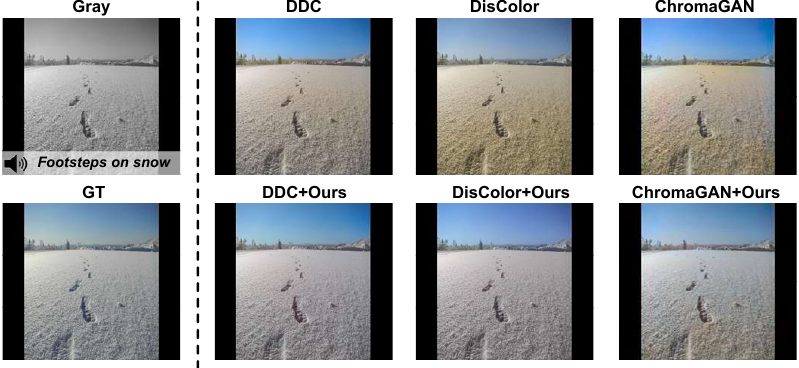}
\caption{Comparisons with existing methods~\cite{38,36,29}, which demonstrates that audio can improve the semantic accuracy of the generated colors so that the overall effect matches the real scene situation.}
\label{fig:teaser}
\end{figure}

In order to avoid difficult color semantic inference, many semi-automatic colorization methods \cite{20,23} mainly rely on human interactions, e.g., color scribbles \cite{20}, reference images \cite{23}, to obtain satisfactory results from given color hints. However, these interactive methods are inefficient, labor-intensive, and sensitive to false prompts. With advances in deep learning, a large number of data-driven automatic colorization methods \cite{27,34,32,29,28} have emerged. Based on large-scale datasets such as ImageNet, some scholars attempt to learn a direct mapping from grayscale images to color images by cleverly designing loss functions \cite{29} or introducing external priors \cite{32}.
Moreover, Kang et al. \cite{38} recently introduce a query-based transformer and multi-scale design to generate vivid color images.
Although these algorithms have achieved remarkable results, reasonable coloring is still difficult, especially when the input grayscale image contains few contextual cues related to the scene. As shown in Fig.~\ref{fig:teaser}, the content of the input image is walking on snow, but existing methods cannot reproduce reasonable colors from the single visual modality of the grayscale scene.

To address this problem, from the perspective of scene perception \cite{8}, we thought of introducing the audio modality for visual semantic complementation and enhancement. In many real-world scenarios, especially in early grayscale old films, videos always have accompanied corresponding audio signals, which record the multi-modal information of the same scene. 
In fact, there exist natural scene semantic links between audio and vision. For example, in our daily life, the sound of raindrops tells us that the sky is gloomy, and the crowing of a rooster brings to our mind the image of a rooster with a red crown. Based on these observations, some intersection studies on audiovisual multimodality have been conducted, e.g., audio-assisted classification \cite{41,zhou2023contrastive,zhou2021positive}, semantic segmentation \cite{49,zhou2023audio}, and scene parsing \cite{48,zhou2024vaplan,zhou2024label}. These studies indicate that audio is very helpful to the understanding of visual scenes, which is exactly necessary for the difficult automatic colorization task.

Therefore, we examine the use of scene semantics provided by audio to assist in image colorization, a topic that has not been explored before. A straightforward way is to design a dual-stream network that directly fuses audio and vision features during end-to-end training. However, due to the modal heterogeneity between audio and vision, the visual backbone usually ignores the role of audio semantics, which is also observed in \cite{9}. 
To solve this problem, taking inspiration from reference-based methods \cite{21,23}, the scene semantics of color images can be used as an intermediate bridge for audio-guided colorization.
Specifically, we first pretrain a semantic-guided colorization network to learn the relationship between color and scene semantics, in which a CNN-based network is used to obtain scene semantic features from color images. Then, the visual features are used as supervision of corresponding audio features to obtain the implicit color semantic representations of the audio scene. Finally, the audio semantic representations are fed into the pretrained visual colorization network to achieve audio-infused automatic image colorization (AIAIC). As shown in Fig.~\ref{fig:teaser}, the proposed method rendered the generated colors more realistically. 

Our contributions are summarized as follows:
\begin{itemize}
\item To the best of our knowledge, this is the first study to adopt cross-modality audio information to assist in the image colorization task.
\item A novel audio-infused colorization method is proposed, which enables the network to learn the latent scene color semantics of audio in a self-supervised manner, providing reasonable and effective guidance for visual colorization.
\item The proposed AIAIC method has pluggability. Experimental results demonstrate that incorporating corresponding audio can enhance the performance of existing visual networks.
\end{itemize}

\section{Related Work}
\subsection{Semi-Automatic Colorization}
Due to the uncertainty of image colorization, traditional methods mainly use human interaction—for example, user scribbles \cite{17,18,20}, and reference images \cite{21,22,23,56}—to guide the colorization process, which can be viewed as semi-automatic colorization. Early scribble-based methods \cite{17} propagate color from user-provided hints to the entire image via an optimization approach, whereas learning-based methods \cite{18} additionally introduce a deep prior from a large-scale image dataset. To address the problem of color incompleteness caused by inefficient network design, Yun et al. \cite{20} recently use a vision transformer to selectively color relevant regions. Although these methods have achieved remarkable results, they require too much manual work, and the quality of results is influenced by user preferences. By contrast, reference-based methods \cite{21,22,23} can reduce intensive user efforts. They convey color information by finding the semantic correspondence between the reference and input images, but they require the two images to be highly correlated. In addition, Varun et al. \cite{24} first introduce a new task of colorization from text descriptions. In order to solve the problem of color-object coupling, Weng et al. \cite{25} construct the color-object correlation matrix in the description and the link between text and object regions to achieve accurate color transfer. Different from them, Bahng et al. \cite{26} try to map the text to the palette first.

\subsection{Fully Automatic Colorization}

Fully automatic colorization \cite{27,28,29,30,31,32,33,34} does not require human intervention. They learn semantic information from large-scale image datasets to convert grayscale images directly into plausible colorful images. Using handcrafted features, Cheng et al. \cite{27} first adopt a neural network to colorize images. However, their network architecture is relatively small. Zhang et al. \cite{28} treat colorization as a classification problem and use cross-channel encoding and class rebalancing techniques in the training stage to yield results with diverse and saturated colors. To obtain better semantic representations, a category prior is introduced  to learn global information \cite{5,29}. Similarly, some methods \cite{30,31} use a two-branch architecture to jointly learn pixel embedding and local information, e.g., segmentation or saliency maps. 
In addition, some scholars \cite{34} have attempted to utilize generative color priors to assist in colorization. 
In order to ensure consistency within the same semantic region, Xia et al. \cite{36} introduce superpixel segmentation networks to color from anchors. Furthermore, Weng et al. \cite{CT2} pre-build a luminance selection module with color probability distribution of the dataset. However, this approach relies on manually calculated priors, which is not conducive to generalization. To this end, recently, Kang et al. \cite{38} utilize a query-based transformer to learn semantic-aware color queries. Although these methods have achieved impressive results, generating colors that reasonably match real scenes remains challenging. 
To alleviate this issue, we first attempt to introduce relevant audio to enhance scene understanding, thereby improving colorization performance.

\section{Proposed Approach}
\subsection{Problem Formulation}
Given an input grayscale image $X\in \mathbb{R}^{H\times W\times1}$, the colorization task aims to find a function $Y=\mathcal{F}\left(X\right),Y\in \mathbb{R}^{H\times W\times3}$ to transform the grayscale image $X$ into a colorized image $Y$, where $H,\ W$ are the height and width of the image, respectively.

If the grayscale image is extracted from a video, such as in the case of restoring colors to old movies, we could also obtain the corresponding sound signal, which records extra audio scene information at the same time. Our goal is to utilize the accompanying audio information to enhance the semantic understanding of the scene, thus improving the colorization performance. Therefore, in this study, the input is $\left\{\left(X_i^L,A_i\right)|i=1,\ldots,n\right\}$, where $A_i$ is the audio signal corresponding to $X_i^L$. The whole process is performed in the CIE $\textbf{Lab}$ color space and can be described as follows:
\begin{equation}
  \widehat{Y}^{ab}=\mathcal{F}\left(X^L|A\right),
\label{Eq1}
\end{equation}
i.e., with the aid of audio, the input grayscale image X is mapped from the luminance channel $\textbf{L}$ to its associated color $\textbf{ab}$ channels. $X^L$ denotes the input image under the L luminance channel.

The core problem of this study is how to effectively extract and apply audio semantics to the colorization task. 
Considering the modal heterogeneity and the choice of the network structure for the potential space of each modality \cite{9,10},
we first try to establish the relationship between color reasoning and scene semantics, and then learn the correlation between scene semantics and corresponding audio features.

\begin{figure*}[!t]
  \centering
  \includegraphics[width=\linewidth]{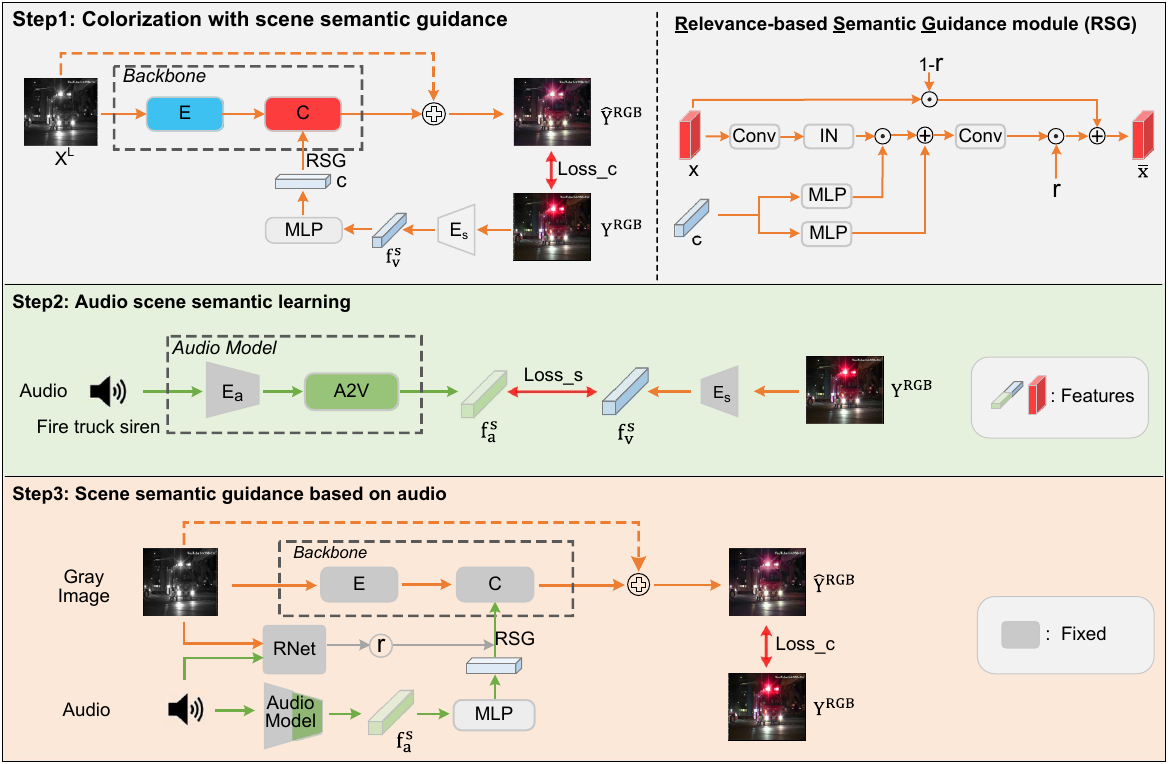}
  \caption{The framework of our proposed method for audio-infused automatic image colorization (AIAIC), which is composed of three steps.}
  \label{FIG2}
\end{figure*}

The overall training process is shown in Fig.~\ref{FIG2}, which can be divided into three steps. We will introduce the details of each step in the following sections.

\subsection{Colorization with Scene Semantic Guidance}
As illustrated in Fig.~\ref{FIG2}, in step 1, we directly use the ground truth color image corresponding to the input grayscale image as auxiliary information to provide scene semantics.

The backbone of the colorization network usually contains two parts, i.e., a feature extraction encoder $E(\cdot)$ and a color generation module $C(\cdot)$. The predicted the missing color information $\widehat{Y}^{ab}$ can be calculated as,
\begin{equation}
  \widehat{Y}^{ab}=C\left(E\left(X^l\right)\right)
\label{Eq2}
\end{equation}
Then, we can obtain the colorized output $\widehat{Y}^{rgb}$ by concatenating $\widehat{Y}^{ab}$ with the input grayscale channel $X^l$ and performing affine transformation.

In this step, we tend to enforce the colorization network to learn the relationship between color reasoning and scene semantics. 
Hence, a CNN-based network is adopted as a semantic feature extraction module $E_s(\cdot)$. The normalized scene color semantics is described as follows,
\begin{equation}
f_v^s=\frac{E_s\left(Y^{rgb}\right)}{{||E_s\left(Y^{rgb}\right)||}_2}
\label{Eq3}
\end{equation}
where $Y^{rgb}$ denotes the ground truth color image, and ${f}_v^s\in \mathbb{R}^{d}$ represents the semantics extracted from a color image. $d$ denotes the feature dimension. Note that ground truth $Y^{rgb}$ is only introduced in the training phase.

Then, after a multi-layer perceptron (MLP), the ${f}_v^s$ are embedded into the color generation module $C(\cdot)$, as:
\begin{equation}
\widehat{Y}^{ab}=C\left(SG(x,c)\right)
\label{Eq4}
\end{equation}
where ${SG}(\cdot)$ denotes a semantic guidance injection module and $c=MLP({\ f}_v^s)$. $x\in \mathbb{R}^{H\times W\times C}$ represents the feature map in the module $C(\cdot)$. 
Notably, as shown in Fig.~\ref{FIG2}, the $DSG(\cdot)$ module should be used here in Eq.~\ref{Eq4}. The use of $SG(\cdot)$ module in the above description is for ease of reading. We will describe the application of $DSG(\cdot)$ module in this context in Sec. 3.5.

Motivated by style transfer methods \cite{11}, which transfer the style of the reference image to the target image, we treat the color semantics $f_v^s$ as color style and then introduce the adaptive instance normalization (AdaIN) \cite{11}
to effectively inject the color semantics. As a result, the AdaIN-based semantic guidance injection module $SG(\cdot)$ is computed as,
\begin{equation}
SG(x,c)=\gamma\left(c\right)\left(\frac{x-\mu\left(x\right)}{\sigma\left(x\right)}\right)+\beta\left(c\right)
\label{Eq5}
\end{equation}
where $\gamma$ and $\beta$ are two MLPs composed of two fully connected (FC) layers and $\mu$, $\sigma$ denote the mean and variance. 

\textbf{Training.} In this step, we use the same color loss $\mathcal{L}_c$ as in the visual baselines~\cite{29,36,38} adopted in this paper.

\subsection{Audio Scene Semantic Learning}
In the previous step, the proposed method learns the relationship between scene semantics and colorization in the same visual modality, instead of establishing difficult cross-modal correspondence between audio and colors. 

For the latter, in this step 2, the scene semantics extracted from color images can be used to supervise the semantics extraction from corresponding audios.

As shown in Fig.~\ref{FIG2}, given the audio signal $A$, the audio feature $f_a$ is firstly obtained by a sound encoder $E_a(\cdot)$. After that, we map $f_a$ to the visual feature space through a projection module $A2V(\cdot)$ constructed by several FC layers to yield the latent semantic feature $f_a^s\in \mathbb{R}^{d}$. The process can be expressed as:
\begin{equation}
f_a^s=A2V\left(E_a\left(A\right)\right)
\label{Eq6}
\end{equation}

\textbf{Training}. The following loss function is used for optimization to enable learning the latent scene semantics of audio: 
\begin{equation}
    \mathcal{L}_s= {{\left \| f_a^s -f_v^s \right \|}}_{2}^{2} 
\label{Eq7}
\end{equation}

Owing to the well-designed multistep training strategy, the audio semantic extraction process is constrained by the scene semantics extracted from color images, which can get rid of the dependence on manual labels of audio semantics. 

\subsection{Scene Semantic Guidance Based on Audio}
Assuming that $f_a^s$ has learned the scene color information from audio, then it can replace $f_v^s$ and be plugged into the previously pre-trained colorization network in step 1.

\textbf{Training}. Considering that the semantic projection module $A2V(\cdot)$ might be suboptimal for colorization, we continue fine-tuning it in the whole network, as shown in Fig.~\ref{FIG2}. Note that the parameters of the colorization backbone are fixed. The color loss $\mathcal{L}_c$ 
continues to be used to further refine the audio scene semantics.

\textbf{Inference:} It should be noted that the step 1 and 2 are only implemented in the training stage. After the three-step training process, the proposed AIAIC network in step 3 can effectively extract and utilize the audio scene semantics to automatically improve scene understanding and coloring accuracy.

\subsection{Dynamic Semantic Guidance Module}
Considering that in real-world scenarios, inconsistencies in audio and visual semantic content, as well as instances of audio absence, are sometimes encountered, we incorporate a modal relevance mechanism in SG module, i.e.,
\begin{equation}
DSG(x,c)=r\odot\left ( SG\left ( x,c \right )  \right ) + \left ( 1-r \right )\odot x 
\label{Eq9}
\end{equation}

This mechanism enables the model to adaptively enhance colorization results according to the correlation between the audio and the visual scene, while ensuring that the visual backbone remains usable when audio is missing.

Next, we will elaborate on the relevance mechanism under conditions with audio and without audio.

\textbf{1) Without audio.} 
When the audio signal is corrupted and inaccessible, $r$ is directly set to 0, i.e., the AIAIC network degenerates to coloring in the visual unimodality. To ensure the standalone capability of the visual backbone in this case, we employ this mechanism beforehand in step 1, i.e., $DSG(\cdot)$ is used in Eq.~\ref{Eq4}, where we mask some ground truth $Y^{rgb}$ inputs. This operation allows the pre-trained colorization network to adapt to the situation where auxiliary branches are absent, thereby enhancing the robustness of the subsequent audio-infused colorization network.

\textbf{2) With audio.}
When audio is available, considering the existence of irrelevant audio-visual scenes, e.g., voice-over and background music, a relevance network is designed to derive the relevance $r\in \left ( 0,1 \right ) $ between audio and vision. 
\begin{figure*}[!t]
  \centering
  \includegraphics[width=0.7\linewidth]{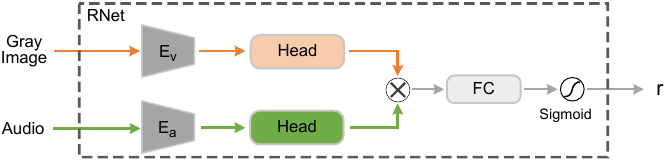}
  \caption{The framework of designed relevance network (RNet).}
  \label{FIG3}
\end{figure*}
As shown in Fig.~\ref{FIG3}, we utilize the pre-trained encoders and trainable heads to extract the features $f_a^r\in \mathbb{R}^{d'}$ and $f_v^r\in \mathbb{R}^{d'}$ from the input audio and image, respectively. 
Subsequently, we compute their cosine similarity and utilize a FC layer followed by a Sigmoid function to map this similarity score to the range $(0,1)$, yielding the relevance $r$. This process can be formulated as follows:
\begin{equation}
    f_{av}^r=\frac{f_{a}^r}{\left \| f_{a}^r \right \| } \otimes \left ( \frac{f_{v}^r}{\left \| f_{v}^r \right \| } \right )^{T}
\label{Eq11}
\end{equation}
\begin{equation}
    r = Sigmoid\left ( f_{av}^r\cdot W \right)
\label{Eq11-2}
\end{equation}
where $\otimes$ denotes the matrix multiplication and $W$ is the learnable parameter. 
For training the relevance network, we view it as a binary classification task, and use the binary cross-entropy (BCE) loss  for optimization:
\begin{equation}
    \mathcal{L}_{r} = BCE(r,h)
    \label{Eq12}
\end{equation}
where $h=1$ or $0$ denotes audio and
vision are relevant or irrelevant, respectively. In the training phase, irrelevant audio-visual pairs are constructed by randomly sampling audio from different videos.

\section{Experiments}
\subsection{Experimental Setting}
\textbf{Datasets.} 
Considering that there is no publicly available dataset containing audio in the filed of colorization, we perform experiments on two existing audiovisual datasets.
\begin{table*}[!t]
  \caption{Quantitative results between our method and three visual baselines.}
  \label{tab:1}
  \resizebox{\textwidth}{!}{
\begin{tabular}{cccccccccc}
\toprule
\multirow{2}{*}{Methods} & \multicolumn{3}{c}{VGGSound}                                       & \multicolumn{3}{c}{AVE}                                             & \multicolumn{3}{c}{VGGSound\_OOD}                                                \\   \cmidrule(lr){2-4} \cmidrule(lr){5-7}  \cmidrule(lr){8-10}
                        &\textit{LIPIS}\textbf{↓}& \textit{PSNR}\textbf{↑}& \textit{SSIM}\textbf{↑}                & \textit{LIPIS}\textbf{↓}& \textit{PSNR}\textbf{↑}& \textit{SSIM}\textbf{↑}                   & \textit{LIPIS}\textbf{↓}& \textit{PSNR}\textbf{↑}& \textit{SSIM}\textbf{↑}                \\  \midrule
DisColor                & 0.156          & 24.195          & 0.937          & 0.145          & 25.224          & 0.943          & 0.161          & 23.881          & 0.937                  \\
\textbf{DisColor+Ours}     & \textbf{0.147} & \textbf{24.463} & \textbf{0.938} & \textbf{0.136} & \textbf{25.475} & \textbf{0.946} & \textbf{0.147} & \textbf{24.620} & \textbf{0.942} \\ \midrule
DDC                     & 0.145          & 23.908          & 0.925          & 0.132          & 24.991          & 0.933          & 0.152          & 23.535          & 0.926                   \\
\textbf{DDC+Ours}       & \textbf{0.138} & \textbf{24.446} & \textbf{0.936} & \textbf{0.128} & \textbf{25.315} & \textbf{0.941} & \textbf{0.147} & \textbf{23.944} & \textbf{0.935}  \\ \midrule
ChromaGAN               & 0.153          & 24.381          & 0.922          & 0.142          & 25.070          & 0.925          & 0.158          & 24.149          & 0.923                   \\
\textbf{ChromaGAN+Ours} & \textbf{0.152} & \textbf{24.783} & \textbf{0.924} & \textbf{0.138} & \textbf{25.812} & \textbf{0.933} & \textbf{0.154} & \textbf{24.940} & \textbf{0.932}  \\ \bottomrule
\end{tabular}
}
\end{table*}
\begin{figure*}[h]
  \centering
  \includegraphics[width=1\linewidth]{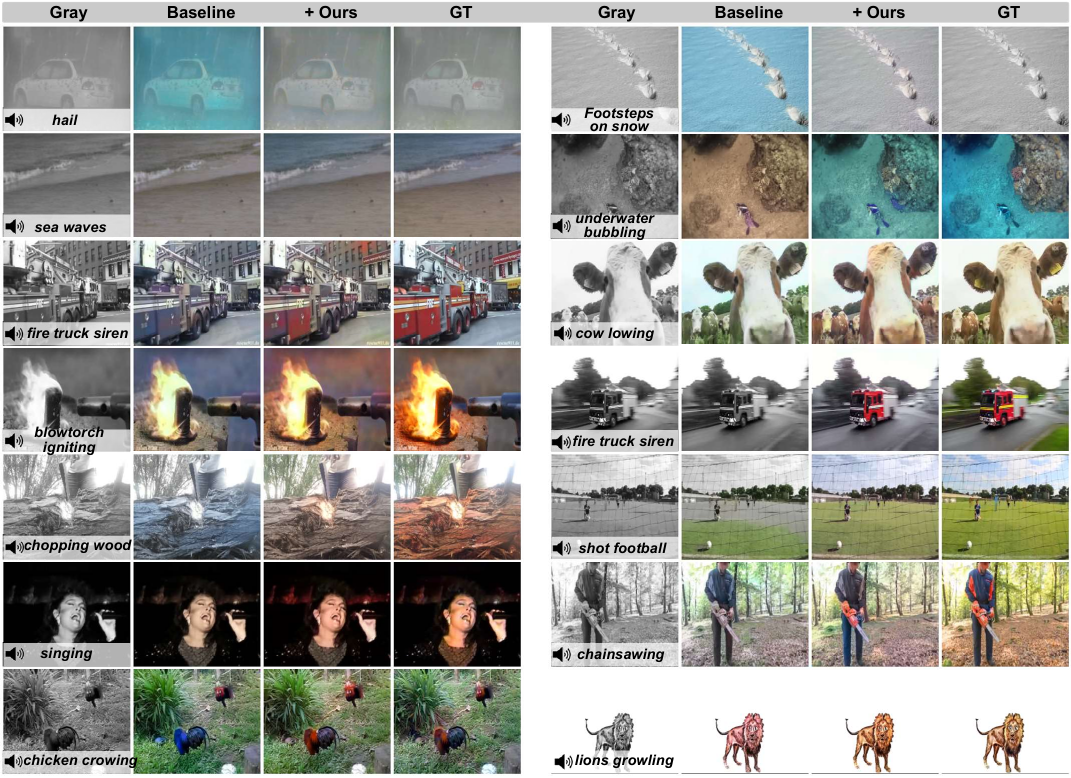}
  \caption{Visual comparisons with the baselines. Our proposed AIAIC method can generate colors that better conform to the actual scene, e.g., sea wave and diving (second row), while enhancing the colors of the subjects in the scene, such as flame (fourth row) and lion (last row).}
  \label{FIG5}
\end{figure*}

\textit{VGGSound}~\cite{13}: VGGSound is a large-scale audiovisual dataset comprising 220,000 10-second videos across 300 distinct sound categories. In this dataset, the objects that emit sound are visible, i.e., the audio and vision are synchronized in content. This feature is particularly conducive for investigating the auditory influence on visual colorization. We utilize a subset encompassing 164 categories (e.g., dog barking, skiing, talking, chicken crowing, playing violin, diving) for training and validation. Each video is sampled at 1 frame per second, with the middle frame and audio selected to form audio-image pairs. The resulting training and validation sets consist of 77,704 and 6,548 pairs, respectively.

\textit{AVE}~\cite{51}: AVE dataset consists of 402 10-second videos for testing. Most of their sound categories overlap with the aforementioned 164 categories. Unlike VGGSound, each video within AVE may include some asynchronous audio-visual segments. For each 1-second segment, we extract the middle frame along with its corresponding audio. Eventually, 4,020 pairs are formed for validation.

Furthermore, to evaluate our performance in unknown audio-visual scenarios, we randomly select additional 1,000 pairs from the original VGGSound dataset to form VGGSound\_OOD. This subset encompasses 20 sound categories that are excluded from the above 164 categories.

\textbf{Implementation details.} To validate the effectiveness and pluggability of the proposed method, we employ three visual-only SOTA baselines: ChromaGAN~\cite{29}, DisColor~\cite{36}, and DDC~\cite{38}, as our colorization backbones. $E(\cdot)$ and $C(\cdot)$ are initialized with pretrained weights provided by respective method. Regarding the position of the DSG module, for ChromaGAN, it is inserted before the penultimate seventh layer of its coloring network. For DisColor, the DSG module is incorporated after the first convolutional layer in the refine net of its coloring module. For DDC, the DSG module is added following the first convolutional layer of its decoder. Furthermore, ResNet-18~\cite{13} pre-trained on VGGSound and VGG-19~\cite{52} pre-trained on ImageNet are used as $E_a(\cdot)$ and $E_v(\cdot)$, respectively. 
The visual scene semantic extraction module $E_s(\cdot)$ comprises four convolutional blocks and two linear layers. Each convolutional block consists of a Convolutional layer, a ReLU function, and a Pooling layer. 

In the training stage, for step 1 and step 3, the network is trained for 20 and 10 epochs with a batch size of 16, respectively. The color loss $\mathcal{L}_c$ and optimizer remain the same as those used in visual backbones. In step 2, we utilize Adam optimizer with an initial learning rate of 0.001 and conduct training for 20 epochs using a batch size of 64. To ensure fairness, we also fine-tune all baselines on the VGGSound training set using their respective pretrained weights. In the inference stage, we use 3 quantitative metrics to measure the colorization results, including Learned Perceptual Image Patch Similarity (LPIPS), Peak Signal-to-Noise Ratio (PSNR), and Structural Similarity Index (SSIM).

\subsection{Results of Colorization}

\textbf{Quantitative results.} 
After training on the VGGSound training set, we evaluate all methods directly across all validation sets. As shown by the metrics in Table~\ref{tab:1}, incorporating audio improves the colorization performance and makes the generated colors more similar to the color of the original image. Moreover, our proposed method shows significant improvement over the baseline methods in unknown audiovisual scenarios, revealing that our approach has certain generalizability.

\textbf{Qualitative results.} To more intuitively demonstrate the effectiveness of our method, we give some visual comparisons in Fig.~\ref{FIG5}. It can be found that by relying only on a single visual modality, the existing visual models are sometimes unable to obtain the correct semantic information of the scene, making the generated color not match the actual situation. For example, for the snow image in the first row, the visual baseline tend to yield a blue color owing to fewer contextual clues. In fact, this grayscale image corresponds to the scene of walking on snow. When we inject the counterpart sound, we can find that it can complement the scene knowledge for the model and correct the generated color. The same is true for the diving scene and the sea wave scene in the second row. For images in which the overall color is not distinct, the associated sound could still enhance the color of the subjects in the scene, such as the color depth of flame in the fourth row. 
\begin{table*}[!t]
\caption{Quantitative comparison of the ablation experiments. Bold represents the best and underline represents the second.}
\centering
 \label{tab:2}
 \resizebox{0.85\textwidth}{!}{
\begin{tabular}{cccccccc}
\toprule
 \multirow{2}{*}{Settings} & \multicolumn{3}{c}{VGGSound}                      & \multicolumn{3}{c}{AVE}      \\ \cmidrule(l{7pt}r{4pt}){2-4} \cmidrule(l{7pt}r{4pt}){5-7} 
                                   & \textit{LIPIS}\textbf{↓}& \textit{PSNR}\textbf{↑}& \textit{SSIM}\textbf{↑}    &  \textit{LIPIS}\textbf{↓}& \textit{PSNR}\textbf{↑}& \textit{SSIM}\textbf{↑}      \\ \midrule
\textbf{full (DisColor-based)}                                  & \textbf{0.147} & \textbf{24.463} & \textbf{0.938} & \textbf{0.136} & \textbf{25.475}    & \textbf{0.946} \\
                                 w/o multi-step                                   & 0.163          & 23.752          & 0.929          & 0.151          & 24.583          & 0.932           \\
                                 w/o $r$                                       & {\ul 0.149}          & {\ul 24.435}          & 0.936         & {\ul 0.140}          & {\ul 25.374} & 0.942                    \\
                                 w/o $r$ (missing audio)                            & 0.807          & 7.467           & 0.103          & 0.792          & 7.670           & 0.098                              \\
                                 full (missing audio)                                &  0.154          & 23.970          & {\ul 0.938}          & 0.143          & 24.822          & {\ul 0.943} 
                                 \\ \midrule
\textbf{full (DDC-based)}                                  & \textbf{0.138} & \textbf{24.446} & \textbf{0.936} & \textbf{0.128} & \textbf{25.315}    & \textbf{0.941}  \\
                                  w/o multi-step                                   & 0.144          & 24.200          & 0.930          & 0.131          & 25.128 & 0.938                    \\
                                 w/o $r$                                       & 0.143          & 24.174          & 0.929          & 0.131          & 25.089          & 0.933                   \\
                                 w/o $r$ (missing audio)                            & 0.201          & {\ul 23.396}          & 0.890          & 0.179          & 24.225          & 0.889                    \\
                                 full (missing audio)                                & {\ul 0.141}          & 24.221          & {\ul 0.931}          & {\ul 0.129}          & {\ul 25.208}          & {\ul 0.939}  
              \\ \midrule
\textbf{full (ChromaGAN-based)}                                  & {\ul 0.152}    & \textbf{24.783} & \textbf{0.924}    & {\ul 0.138} & \textbf{25.812} & \textbf{0.933}     \\
                                 w/o multi-step                                   & 0.153          & 24.471          & 0.922          & \textbf{0.137}          & 25.235          & 0.927                    \\
                                 w/o $r$                                       & \textbf{0.150} & {\ul 24.761}          & {\ul 0.923} & 0.139          & {\ul 25.674}          & {\ul 0.931}           \\
                                 w/o $r$ (missing audio)                            & 0.639          & 10.546          & 0.311          & 0.628          & 10.757          & 0.292                    \\
                                 full (missing audio)                                & 0.157          & 24.490          & 0.920          & 0.142          & 25.319          & 0.923     
                                 \\ \bottomrule
\end{tabular}
}
\end{table*}
\subsection{Ablation Study}
\begin{figure}[!t]
  \centering
  \includegraphics[width=\linewidth]{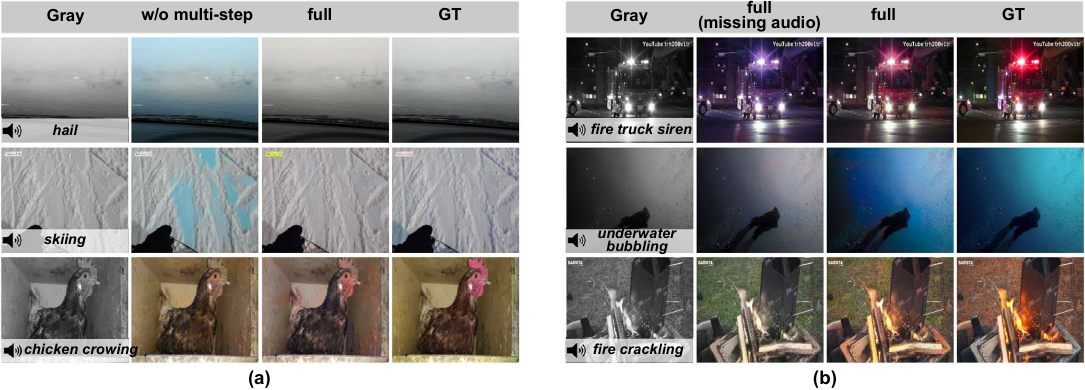}
  \caption{Qualitative comparisons for demonstrating that the incorporation of
audio and multi-step training strategy can effectively complement and enhance the
scene semantic understanding for the visual model to generate more accurate
colors. 
  }
  \label{FIG6}
\end{figure}

\textbf{Effectiveness of audio.} 
To further explore the effectiveness of audio, we directly exclude audio by setting r to 0 in Eq.~\ref{Eq9}. As illustrated in Table~\ref{tab:2}, comparing `full’ and `full (missing audio)’, we observe a noticeable performance decline when audio is omitted, which shows that audio can
effectively improve colorization performance.
Additionally, Fig.~\ref{FIG6} (b) provides some qualitative comparisons. It can be found that the addition of relevant audio leads to a better understanding of scene; for instance, blue for diving, red for ambulance.

\textbf{Effectiveness of multi-step training.} 
The purpose of the multi-step training is to learn the implicit scene color semantics of the audio, thus providing an effective aid for visual coloring. If we incorporate audio directly into the visual model for end-to-end training, i.e., ‘w/o multi-step’, due to the modal heterogeneity between audio and vision, the model usually ignores the role of audio and cannot successfully establish the correspondence between audio and colorization. Fig.~\ref{FIG6} (a) shows that the colors of the sky and snow are completely incorrect, which demonstrates the importance of scene semantic learning of audio.

\textbf{Effectiveness of Relevance Mechanism (RM) in the DSG module.} 
The RM is designed to enhance the robustness of the AIAIC model. Specifically, it can increase dependency on the visual backbone for colorization when audio and vision are irrelevant. Moreover, when audio is not available, degradation to a visual-only model can still allow for basic colorization. To validate these, we conduct two corresponding ablation experiments. 1) Comparison between `full' and `w/o $r$' settings in Table~\ref{tab:2} demonstrates that incorporating the RM 
generally improves performance, especially on AVE dataset containing irrelevant segments. 
2) Furthermore, when audio is unavailable, i.e., `w/o $r$ (missing audio)', colorization fails entirely, as indicated by significant discrepancies in the LIPIS and SSIM metrics compared to the `full' setting. Conversely, in the setting `full (missing audio)’, adding this mechanism allows the model to sustain a certain level of colorization effect as in the visual baseline.

\section{Conclusion}
This paper proposes a novel and pluggable audio-infused automatic image colorization method for the first time, which can use corresponding audio information to enhance the scene semantics and improve the colorization performance. The network is trained in three steps without manual labels of audio semantics. First, the colorization backbone is pretrained with scene semantics extracted from the visual domain. Then, the optimized visual scene semantics are adopted to constrain the learning of audio semantics. Finally, the audio semantics are used to improve the coloring process. Experimental results demonstrate the effectiveness of our proposed audio-guided method.

\subsubsection{\ackname} This work is supported by the National Natural Science Foundation of China under Grant 61972127, Grant 61972129.

\bibliographystyle{splncs04}
\bibliography{0800}

\end{document}